\documentclass[pdflatex,sn-vancouver-num]{sn-jnl}

\usepackage{graphicx}%
\usepackage{multirow}
\usepackage{amsmath,amssymb,amsfonts}%
\usepackage{amsthm}%
\usepackage{mathrsfs}%
\usepackage[title]{appendix}%
\usepackage{xcolor}%
\usepackage{textcomp}%
\usepackage{manyfoot}%
\usepackage{booktabs}%
\usepackage{algorithm}%
\usepackage{algorithmicx}%
\usepackage{algpseudocode}%
\usepackage{listings}%

\theoremstyle{thmstyleone}%
\theoremstyle{thmstyletwo}%

\theoremstyle{thmstylethree}%

\begin{document}

\title[Article Title]{Explainable Counterfactual Reasoning in Depression Medication Selection at Multi-Levels (Personalized and Population)}


\author[1]{\fnm{Xinyu} \sur{Qin}}
\author[3]{\fnm{Mark H.} \sur{Chignell}}
\author[4]{\fnm{Alexandria} \sur{Greifenberger}}
\author[5]{\fnm{Sachinthya} \sur{Lokuge}}
\author[4]{\fnm{Elssa} \sur{Toumeh}}
\author[4]{\fnm{Tia} \sur{Sternat}}
\author[4]{\fnm{Martin} \sur{Katzman}}
\author*[1, 2]{\fnm{Lu} \sur{Wang}}\email{lwang71@central.uh.edu}

\affil[1]{\orgdiv{Department of Biomedical Engineering, Cullen College of Engineering}, \orgname{University of Houston}, \orgaddress{\street{4226 Martin Luther King Boulevard}, \city{Houston}, \postcode{77204}, \state{Texas}, \country{United States}}}
\affil[2]{\orgdiv{Department of Health Systems \& Population Health Sciences, Tilman J. Fertitta Family College of Medicine}, \orgname{University of Houston}, \orgaddress{\street{4226 Martin Luther King Boulevard}, \city{Houston}, \postcode{77204}, \state{Texas}, \country{United States}}}
\affil[3]{\orgdiv{Department of Mechanical \& Industrial Engineering}, \orgname{University of Toronto}, \orgaddress{\street{5 King’s College Road}, \city{Toronto}, \postcode{M5S 3G8}, \state{Ontario}, \country{Canada}}}
\affil[4]{\orgname{START Clinic for Mood and Anxiety Disorders}, \orgaddress{\street{32 Park Road}, \city{Toronto}, \postcode{M4W 2N4}, \state{Ontario}, \country{Canada}}}
\affil[5]{\orgdiv{Department of Psychology}, \orgname{Virginia Tech}, \orgaddress{\street{890 Drillfield Drive}, \city{Blacksburg}, \postcode{24060}, \state{Virginia}, \country{United States}}}




\abstract{\textbf{Background:} This study investigates how variations in Major Depressive Disorder (MDD) symptoms, quantified by the Hamilton Rating Scale for Depression (HAM-D), causally influence the prescription of SSRIs versus SNRIs.  
\textbf{Methods:} We applied explainable counterfactual reasoning with counterfactual explanations (CFs) to assess the impact of specific symptom changes on antidepressant choice.  
\textbf{Results:} Among 17 binary classifiers, Random Forest achieved highest performance (accuracy, F1, precision, recall, ROC-AUC near 0.85). Sample-based CFs revealed both local and global feature importance of individual symptoms in medication selection.  
\textbf{Conclusions:} Counterfactual reasoning elucidates which MDD symptoms most strongly drive SSRI versus SNRI selection, enhancing interpretability of AI-based clinical decision support systems. Future work should validate these findings on more diverse cohorts and refine algorithms for clinical deployment.
}




\maketitle

\section{Introduction}
\label{sec:introduction}
Major depressive disorder (MDD) is a severe mental illness that significantly impacts global public health, leading to deterioration in both physical and mental well-being \cite{sinyor2016suicide}. The Hamilton Rating Scale for Depression (HAM-D) is one of the most extensively employed assessment tools to objectively assess the severity of depression. In research settings, it is clinician-administered, and there are multiple versions, i.e., the 7-item version, 17-item version, 21-item version, and 24-item version \cite{carrozzino2020hamilton, mcintyre2002assessing, hamilton1960rating, hamilton1967development}. The 17-item version is the most frequently employed, which is also the version utilized in this paper. The symptoms of depression are listed in TABLE \ref{tab1} \cite{nixon2020bi}. Each symptom or item is scored ranging from 0-4, 0-3 or 0-2. \textcolor{black}{Historically the HAM-D total scores were used to assess depression severity pre- and post-treatment. It is hypothesized that by identifying patient-specific symptoms, a clinician could tailor treatment by drug class. This could provide input for the ultimate goal of the development of a treatment algorithm in clinical utility for clinicians looking to predict which treatment option should be used based on patient presentation \cite{papakostas2007augmentation}.}
\begin{table}[ht]
\centering
\caption{17 items/symptoms of MDD}
\label{tab1}
\setlength{\tabcolsep}{2pt} 
\renewcommand{\arraystretch}{1.2} 
\begin{tabular}{|p{50pt}|p{190pt}|} 
\hline
\scriptsize \textbf{HAM-D Item}  & \textbf{Symptom/Item Description} \\
\hline
\scriptsize HAM-D01 & Depressed mood \\
\hline
\scriptsize HAM-D02 & Feelings of guilt \\
\hline
\scriptsize HAM-D03 & Suicidal thoughts or actions \\
\hline
\scriptsize HAM-D04 & Insomnia-early (sleep onset delay) \\
\hline
\scriptsize HAM-D05 & Insomnia-middle (mid-sleep wakening) \\
\hline
\scriptsize HAM-D06 & Insomnia-late (early morning wakening) \\
\hline
\scriptsize HAM-D07 & Work and activities (assessing pleasure and functioning) \\
\hline
\scriptsize HAM-D08 & Psychomotor retardation (slow movement/speech) \\
\hline
\scriptsize HAM-D09 & Psychomotor agitation (restless, fidgeting, etc.) \\
\hline
\scriptsize HAM-D10 & Psychic anxiety (worry, apprehension, etc.) \\
\hline
\scriptsize HAM-D11 & Somatic anxiety (heart racing, sweating, etc.) \\
\hline
\scriptsize HAM-D12 & Loss of appetite \\
\hline
\scriptsize HAM-D13 & Tiredness/pain \\
\hline
\scriptsize HAM-D14 & Loss of sexual interest \\
\hline
\scriptsize HAM-D15 & Hypochondriasis \\
\hline
\scriptsize HAM-D16 & Weight loss \\
\hline
\scriptsize HAM-D17 & Lack of insight \\
\hline
\end{tabular}
\end{table}

\begin{figure*}[!t]
\centerline{\includegraphics[width=\textwidth]{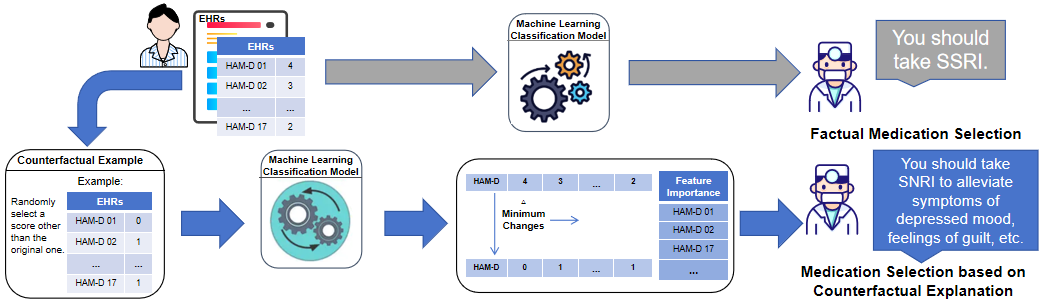}}
\caption{Illustration of explainable counterfactual reasoning on MDD medication selection.}
\label{C1}

\end{figure*}

\textcolor{black}{Distinguishing between different categories of antidepressants based on individual HAM-D item scores can be clinically relevant and beneficial for several reasons. Two commonly used categories of antidepressants as the first-line treatments for MDD are Selective Serotonin Reuptake Inhibitors (SSRIs) and Serotonin-Norepinephrine Reuptake Inhibitors (SNRIs) \cite{kennedy2016canadian}}. SSRIs primarily increase serotonin levels in the synapse, which may be more effective in alleviating mood and anxiety-related symptoms, while SNRIs increase both serotonin and norepinephrine levels in the synapse, potentially offering additional benefits for patients with significant anxiety or physical symptoms like pain \cite{thase2001remission, shelton2019serotonin}. Despite SSRIs and SNRIs being first-line treatments, up to 50\% of patients do not respond adequately to these medications \cite{garcia2012treatment}. This inadequate response may result from difficulties in accurately predicting which treatment will work best for a patient’s specific symptoms. Such challenges often lead to residual symptoms, worsened functioning, more chronic episodes, and increased healthcare costs \cite{garcia2012treatment}. It is hypothesized that artificial intelligence (AI) can aid in identifying preferential response patterns based on individual symptom profiles, helping clinicians make more targeted antidepressant prescribing decisions for MDD patients.

Current AI-based clinical decision support systems (CDSSs) merely leverage the capabilities of the Electronic Health Records (EHRs) system to enhance the healthcare delivery without considering the complexed relationship between multiple symptoms and medication selection in the clinical decision making process. More specifically, AI CDSSs highly rely on the population data to train the model, which may lead to biases if the dataset does not represent all patient groups adequately and definitely ignores patient variabilities at personalized level \textcolor{black}{\cite{gianfrancesco2018potential}}. To better address this issue, \textcolor{black}{we aimed to}
 quantify the relationship between multiple symptoms and medication selection at both personalized and population levels. 

However, AI-based quantification (e.g., feature importance scores that indicate the contribution of each variable to the model’s prediction) can be difficult for clinicians to interpret. Therefore, integrating interpretability into AI CDSSs is crucial to provide clear explanations for predictions. Most machine learning (ML) models, particularly deep learning, are often seen as 'black boxes' due to their opaque decision-making processes, leading to a lack of trust and acceptance. eXplainable Artificial Intelligence (XAI) was introduced to address this issue by making model decisions more transparent \cite{arrieta2020explainable}.

ML prediction models may be used to aid in personalized medication treatment, which can help optimize treatment outcomes. In the context of medical AI, “causality” further  aids interpretability by explaining why an AI model suggests certain treatments for a patient through direct links between input symptoms and output recommendations, making  the system’s reasoning clear and understandable \textcolor{black}{ \cite{holzinger2019causability}.} Counterfactual reasoning, a mode of thinking that considers alternative scenarios and what might have happened under different conditions \textcolor{black}{\cite{roese2014might}}, is particularly effective in clinical contexts to learn the causality. \textcolor{black}{In depression treatment, counterfactual reasoning enables questions e.g., 'Would a patient’s depressive symptoms have improved with a different antidepressant?' This provides deeper insights into the causal effects of medication selection beyond mere correlations \cite{hernan2010causal}.}




Counterfactual explanations (CFs) is a method \textcolor{black}{that}
adopted the concept of counterfactual reasoning in XAI that explains machine learning model predictions by describing how an outcome would change if the input data were different by generating multiple CFs, which provides an awareness of the model's behavior and decision-making process \cite{kim2016examples}. \textcolor{black}{By generating counterfactual examples that represent hypothetical interventions on the input features (HAM-D symptoms), we can assess the causal impact of altering those symptoms on the model's predicted medication selection \cite{wachter2017counterfactual}.} 

Recent advancements have led to various CFs methods, including Feasible and Actionable Counterfactual Explanations \cite{poyiadzi2020face}, Growing Spheres \cite{laugel2017inverse}, and Multi-objective CFs \cite{dandl2020multi}. While these techniques contribute significantly to the field, they often generate either a single counterfactual example or produce relatively homogeneous explanations, limiting the diversity of decision-making options crucial for practical applications. To address these limitations, our study employs the Diverse Counterfactual Explanations (DICE) method \cite{mothilal2020explaining}. DICE enhances the generation of multiple, diverse CFs for a single instance and allows for the imposition of constraints to prevent specific variables from changing. This approach provides more tailored and practical decision-making support, aligning with regulatory requirements for explainable AI and fostering trust between healthcare professionals and the system \cite{ribera2019can}. Fig. \ref{C1} presents a more concrete illustration to show MDD medication selection based on explainable counterfactual reasoning.

\begin{figure}[!t]
\centerline{\includegraphics[width=0.7\columnwidth]{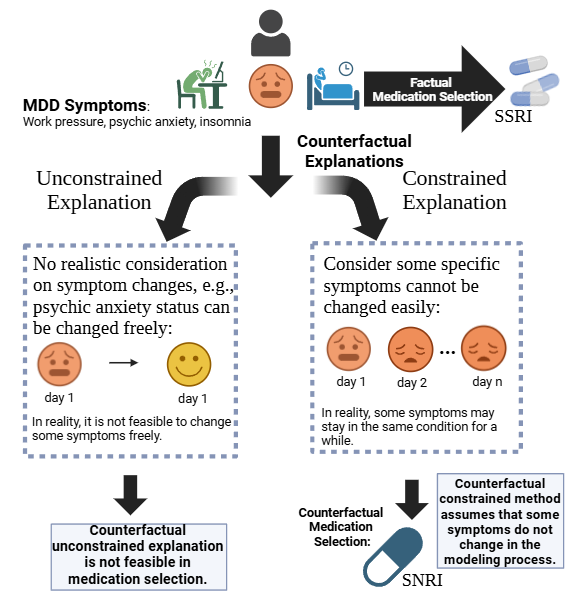}}
\caption{Illustration of constrained explanation and unconstrained counterfactual explanations.}
\label{S1}

\end{figure}

In real-world scenarios, particularly in the medical field, there are numerous constraints generating CFs. For instances, chronic stressors such as work pressure may be observed in patients with MDD and may not feasibly change in a short period of time (see the first panel in Fig. \ref{S1}). The Fig. \ref{S1}  also shows a scenario that does not take real-world factors into account and a scenario that considers real-world factors (see the first column in Fig. \ref{S1}), factoring in the likelihood that a patient typically will not experience immediate symptomatic relief (see the second column in Fig. \ref{S1}).


\begin{figure*}[!t]
\centerline{\includegraphics[width=\textwidth]{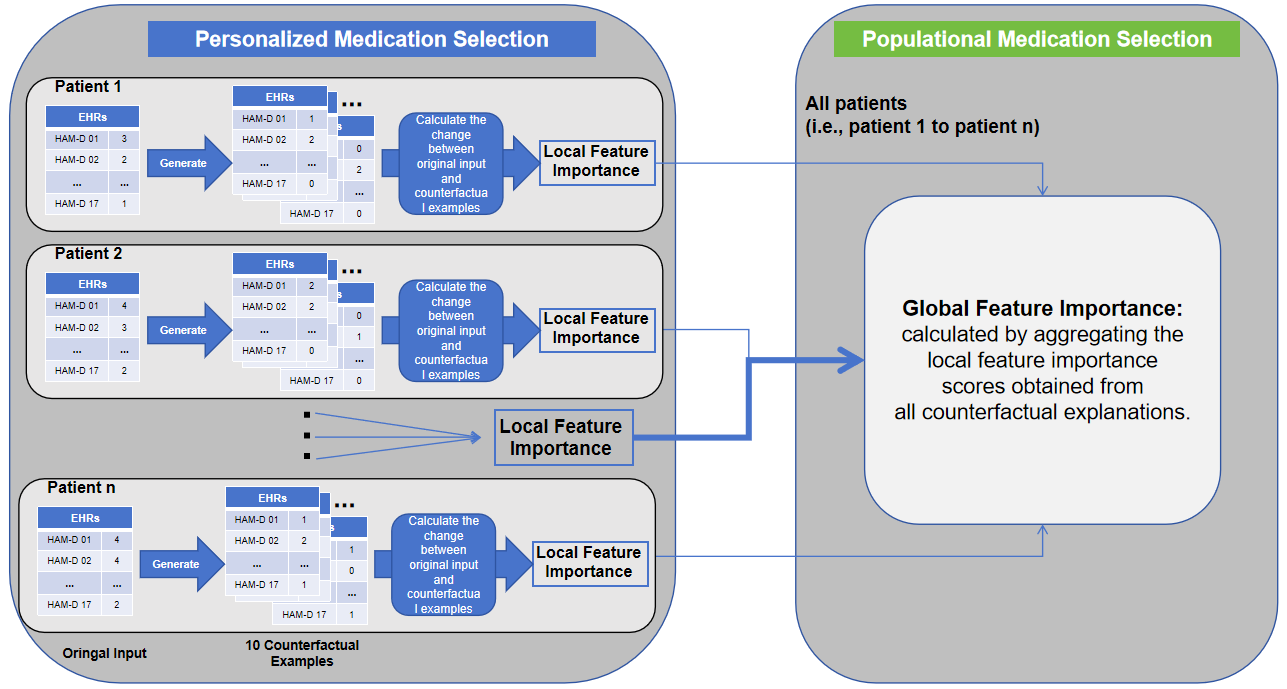}}
\caption{The comparison of local and global feature importance schemes of CFs on personalized and population levels, respectively.}
\label{l1}
\end{figure*}

In XAI, model interpretability has been addressed through various feature importance methods, including local and global approaches. Local feature importance quantifies how individual features influence a specific prediction, with methods like LIME approximating non-linear models using linear ones \cite{ribeiro2016should} and SHapley Additive
exPlanations method providing a unified framework for such calculations \cite{lundberg2017unified}. Global feature importance, on the other hand, measures how features contribute to predictions across the entire dataset by using surrogate models \cite{craven1995extracting, ribeiro2016should}. However, these methods simulate a surrogate model and may not fully reflect the original model's behavior. \textcolor{black}{In addition, they are not ideally suited for the 'what-if' scenarios central to our study. More specifically, they focus on explaining feature importance without providing insight into how altering features would change the model's outcome. In our clinical context, understanding how depressive symptoms change, as evaluated by the HAM-D, based on specific drug selection is crucial, and CFs directly address this by generating alternative scenarios.} 
In CFs, local feature importance of one instance is determined by the frequency of feature changes when generating multiple CFs for a given instance \cite{kommiya2021towards}. The higher the frequency of changing a feature when generating CFs, the more important it is for the model's decision-making process. The global feature importance is computed by summing up local feature importance from all explanations. The comparison of local and global feature importance schemes is illustrated in Fig. \ref{l1}. \textcolor{black}{In summary, CFs not only calculate feature importance, but also visualize these influences, providing clinicians with an intuitive understanding of the reasoning behind medication recommendations \cite{ali2023enlightening}.}

By doing so, our work makes the following contributions: 
\begin{itemize}
    \item We provide quantifiable AI CDSSs to predict medication selections for patients with MDD at both the personalized and population levels.
    \item We establish a method that may in the future support clinicians to offer treatment recommendations based on personalized health conditions for each individual patient.
    \item We develop an explainable AI method on MDD medication selection to improve the trust and intake of AI by human experts, i.e., clinicians in our case.
\end{itemize}

\section{Method}

\textcolor{black}{In the context of our study, the input features $X$ represent an individual patient's various HAM-D item scores, which characterize the severity of different depressive symptoms. The counterfactual example $X'$ refers to a slightly altered version of the patient's original HAM-D item scores. The difference between $X$ and $X'$ captures the changes in symptom severity that would lead to a different predicted antidepressant medication class being selected for the patient.} Specifically, the Eq. (\ref{eq1}) aims to find the counterfactual example $X'$ that minimizes the loss between the target medication class $y_{target}$ and the model's prediction $f(X')$, while also keeping $X'$ close to the original input $X$. \textcolor{black}{This allows us to identify the minimal changes in individual HAM-D item symptom scores that would result in the model recommending a different antidepressant medication class (e.g., switching from an SSRI to an SNRI) for the individual patient.} The formula for generating CFs is expressed as follows from Unconditional Counterfactuals \cite{wachter2017counterfactual}:

\begin{equation}
\label{eq1}
   \min \mathcal{L}(f(X'), y_{target}) + \lambda \cdot dist(X', X). \quad 
\end{equation}

The function $f(\cdot)$ represents a trained target model, which can range from linear models to various machine learning models, such as random forests. \textcolor{black}{This model takes a patient's individual HAM-D item scores as input and predicts the most suitable antidepressant medication.}
The $\mathcal{L}(\cdot)$ is a loss function, typically implemented using the $L_{1}$ loss. The $dist(\cdot)$ denotes the distance function. As HAM-D scores include both continuous and categorical variables, we need distinct definitions for different types of variables:

\begin{equation}
\label{eq2}
   dist\_cont(X', X)= \frac{1}{d_{cont}}\sum_{p=1}^{d_{cont}} \frac{\left\vert X'^{p} - X^{p} \right\vert}{MAD_{p}},
\end{equation}

\begin{equation}
\label{eq3}
   dist\_cat(X', X)= \frac{1}{d_{cat}} \sum_{p=1}^{d_{cat}} \mathbb{I}(X'^{p} \neq X^{p}).
\end{equation}

Given that features in the HAM-D scores may span diverse ranges, computing the median absolute deviation (MAD) for each continuous variable offers a robust measure. The distance function of continuous variables is obtained by Eq. (\ref{eq2}) from \cite{wachter2017counterfactual}. Here, \textcolor{black}{$d_{cont}$ represents the total count of continuous variables in the HAM-D scale, and $MAD_{p}$ denotes the median absolute deviation for the $p^{th}$ HAM-D item.}
For categorical features in the HAM-D scale, a simple distance function is employed as Eq. (\ref{eq3}) from \cite{wachter2017counterfactual}: 1 is assigned if the value of the $X'$ differs from the $X$, otherwise, it is assigned a value of zero. Here, $d_{cat}$ represents the summary of categorical variables in HAM-D, and $\mathbb{I}$ serves as a binary indicator (0/1).
To generate multiple CFs ($X_{1}', \ldots, X_{k}')$ for computing feature importance to assist clinicians' decision-making, we adapt the DICE method. This approach builds upon the work of \cite{wachter2017counterfactual} by introducing a module that enhances the diversity of CFs. Its formula can be expressed as from \cite{mothilal2020explaining}:

\begin{align}
\label{eq4}
    \min \frac{1}{k} \sum_{i=1}^{k} \mathcal{L}(f(X_{i}'), y_{target}) &+ \frac{\lambda_{1}}{k} \sum_{i=1}^{k} dist(X_{i}', X) \\ \nonumber
    &- \lambda_{2}dpp\_diversity(X_{1}', \ldots, X_{k}'),
\end{align}
where $k$ represents the total number of generated CFs, while $\lambda_{1}$ and $\lambda_{2}$ are hyperparameters that can be set manually around 0 - 1. Each $X_{i}'$ is randomly generated within the value range of the utilized HAM-D dataset. $dist(\cdot)$ maintains the same definition as \cite{wachter2017counterfactual}, while $\mathcal{L}(\cdot)$ from \cite{mothilal2020explaining} is expressed as:

\begin{equation}
\label{eq5}
loss = \max\left(0, 1 - z \cdot\left(f\left(X'\right)\right)\right).
\end{equation}

\textcolor{black}{Note that, when $y_{target}$ = 0 (SSRIs), $z$ is -1 and when $y_{target}$ = 1 (SNRIs), $z$ is 1.} Our objective of generating CFs is to make the model's output $f(X')$ exceed or fall below a fixed threshold (usually 0.5), without necessarily requiring it to closely match the expected output $y_{target}$ (0 or 1).
We want to generate a set of $k$ CFs, and they will all lead to a different antidepressant medication class decision than the original input $X$.
The diversity metric, denoted as $dpp\_diversity(\cdot)$, is computed using the determinantal point processes (DPP) technique \cite{kulesza2012determinantal}:

\begin{equation}
\label{eq6}
    dpp\_diversity = \det(K),
\end{equation}
where $K_{i,j} = \frac{1}{1 + dist\left(X_{i}', X_{j}'\right)}$, if the generated CFs $X_{i}'$ and $X_{j}'$ are more similar, the distance between them is smaller, leading to a larger $\det(K)$. Incorporating this metric into the generation process promotes the creation of more diverse CFs, providing clinicians with a wider range of potential and feasible options for adjusting HAM-D scores to switch between SSRIs and SNRIs recommendations.

\textcolor{black}{To generate the CFs, we start with the trained machine learning model and an original patient instance along with its predicted medication class. We then set the target medication class that the counterfactual should be classified as (typically the opposite of the original prediction). Next, we iteratively optimize the counterfactual example by making minimal changes to the patient's original HAM-D item scores, using gradient descent to minimize the loss between the target and predict medication class while also keeping the counterfactual close to the original data distribution.}

\section{Experimental Results}
\label{sec:guidelines}

\subsection{Clinical Trial Depression Dataset}

The data utilized in our paper are provided by Eli Lilly and company and consisted of the compiled and analyzed clinical trial data of duloxetine and its comparator medications (venlafaxine, paroxetine, and placebo), spanning phases II, III, and IV studies, involving a total of 1468 participants. The dataset included multiple pre- and post-assessments of HAM-D scores obtained from 10 randomized clinical trials. Our research primarily focuses on pharmacotherapeutic decision-making, specifically at baseline, denoted as the V1-HAM-D in this paper. \textcolor{black}{As a clinical trial based dataset, it may contain clinical errors. However, this is one of the motivations for implementing CFs to reduce the influences of these errors on model prediction. CFs explore multiple counterfactual scenarios by altering certain variables of the original instance, which offer a potentially correct version of the original instance. Additionally, we calculated global feature importance based on the entire dataset. This calculation helps minimize bias from clinical errors in individual samples.}

\begin{table}[!htbp]
  \centering
  \resizebox{\columnwidth}{!}{%
    \begin{tabular}{|c|l|l|c|}
      \hline
      \textbf{Drugs categorized by} & \textbf{SNRI} & \textbf{SSRI} & \textbf{Data Distribution} \\
      \hline
      Drug &
        \begin{tabular}[c]{@{}l@{}}All Duloxetine,\\ All Venlafaxine\end{tabular} &
        \begin{tabular}[c]{@{}l@{}}All Escitalopram,\\ All Paroxetine,\\ All Fluoxetine\end{tabular} &
        \begin{tabular}[c]{@{}c@{}}SNRIs: 1070\\ SSRIs: 398\end{tabular} \\
      \hline
      Dosing Version 1 &
        \begin{tabular}[c]{@{}l@{}}Venlafaxine $\geq$ 150 mg\\ Paroxetine $\geq$ 50 mg\\ Duloxetine $\geq$ 60 mg\end{tabular} &
        \begin{tabular}[c]{@{}l@{}}All Escitalopram,\\ All Fluoxetine\\ Venlafaxine $<$ 150 mg\\ Paroxetine $<$ 50 mg\\ Duloxetine $<$ 60 mg\end{tabular} &
        \begin{tabular}[c]{@{}c@{}}SNRIs: 930\\ SSRIs: 538\end{tabular} \\
      \hline
      Dosing Version 2 &
        \begin{tabular}[c]{@{}l@{}}Venlafaxine $>$ 150 mg\\ Paroxetine $>$ 50 mg\\ Duloxetine $>$ 60 mg\end{tabular} &
        \begin{tabular}[c]{@{}l@{}}All Escitalopram,\\ All Fluoxetine\\ Venlafaxine $\leq$ 150 mg\\ Paroxetine $\leq$ 50 mg\\ Duloxetine $\leq$ 60 mg\end{tabular} &
        \begin{tabular}[c]{@{}c@{}}SNRIs: 347\\ SSRIs: 1121\end{tabular} \\
      \hline
    \end{tabular}%
  } 
  \caption{Antidepressant Drug Categorization}
  \label{tab:classification}
\end{table}

We categorize the antidepressants into three categories based on medication name and dosage, as detailed in TABLE \ref{tab:classification} based on three distinctly different mechanisms of categorization:

(i). The first method of categorization is based on a priori criteria of marketed categorization of the antidepressants (with Venlafaxine and Duloxetine labeled as belonging to the class of SNRIs, and Escitalopram, Paroxetine, and Fluoxetine labeled as SSRIs).

(ii). For Dosing Version 1, SNRIs were considered to be venlafaxine $\geq 150$ mg, paroxetine $\geq 50$ mg, and duloxetine $\geq 60$ mg. All other doses of venlafaxine, paroxetine, and duloxetine, as well as escitalopram and fluoxetine, were considered as SSRIs.

(iii). For the Dosing Version 2 analysis, SNRIs were considered to be venlafaxine $> 150$ mg, paroxetine $> 50$ mg, and duloxetine $> 60$ mg. All other doses of venlafaxine, paroxetine, and duloxetine, as well as escitalopram and fluoxetine, were considered as SSRIs.

This classification approach is therefore adopted for subsequent analyses. \textcolor{black}{In the subsequent data processing, the Synthetic Minority Over-sampling Technique (SMOTE) is employed to balance the imbalanced dataset \cite{nitesh2002smote}.} One-hot encoding is applied to all HAM-D scores, given that those scores are categorical variables. \textcolor{black}{Due to the variations in mechanisms of action, it is necessary to consider other antidepressant classes to offer maximum value for clinical practice. However, clinical studies are often limited by the specifics of data collection. That is, because this analysis begins with the use of a retrospective dataset, we are limited by what protocols were included in the dataset. }

\subsection{Model Selection}

\begin{table*}[!htbp]
    \caption{The evaluation metrics of 17 machine learning methods evaluated under the Dosing Version 2 categorization. One-hot encoding, oversampling, and 5-fold cross-validation are applied on all 17 models to improve the corresponding model performances. The evaluation metrics for both training and testing sets are presented.}
    \centering
    \resizebox{\textwidth}{!}{%
    \begin{tabular}{|l|c|c|c|c|c|c|c|c|c|c|}
        \hline
        \multirow{2}{*}{Method} & \multicolumn{5}{c|}{\small Training Set} & \multicolumn{5}{c|}{\small Testing Set} \\ \cmidrule(lr){2-6} \cmidrule(lr){7-11}
        & Accuracy & F1 Score & Precision & Recall & ROC-AUC 
        & Accuracy & F1 Score& Precision & Recall & ROC-AUC \\ \hline
        \multicolumn{11}{|c|}{\textbf{Ensemble machine learning models}} \\ \hline
        Random Forest & \textbf{0.9991} & \textbf{0.9991} & \textbf{0.9991} & \textbf{0.9991} & 0.9997 
        & \textbf{0.8497} & \textbf{0.8496} & \textbf{0.8502} & \textbf{0.8497} & \textbf{0.9253} \\
        CatBoost & 0.9537 & 0.9537 & 0.9557 & 0.9537 & 0.9921 
        & 0.8261 & 0.8259 & 0.8285 & 0.8261 & 0.8993 \\
        Stacking & 0.9886 & 0.9886 & 0.9887 & 0.9886 & 0.9989 
        & 0.8238 & 0.8239 & 0.8245 & 0.8238 & 0.8960 \\
        Extra Trees & \textbf{0.9991} & \textbf{0.9991} & \textbf{0.9991} & \textbf{0.9991} & \textbf{1.0000} 
        & 0.8220 & 0.8221 & 0.8232 & 0.8220 & 0.9013 \\
        Hist Gradient Boosting & 0.9705 & 0.9704 & 0.9709 & 0.9705 & 0.9961 
        & 0.8207 & 0.8206 & 0.8228 & 0.8207 & 0.8876 \\
        Voting & 0.9756 & 0.9756 & 0.9757 & 0.9756 & 0.9974 
        & 0.8131 & 0.8130 & 0.8153 & 0.8131 & 0.8939 \\
        AdaBoost & 0.7217 & 0.7215 & 0.7222 & 0.7217 & 0.7856 
        & 0.7154 & 0.7152 & 0.7184 & 0.7154 & 0.7753 \\
        Gradient Boosting & 0.7344 & 0.7334 & 0.7375 & 0.7344 & 0.8105 
        & 0.7003 & 0.6992 & 0.7061 & 0.7003 & 0.7760 \\ \hline
        \multicolumn{11}{|c|}{\textbf{Nonparametric machine learning models}} \\ \hline
        K-Nearest Neighbor & 0.9063 & 0.9061 & 0.9104 & 0.9063 & 0.9734 
        & 0.8077 & 0.8072 & 0.8126 & 0.8077 & 0.8658 \\
        Decision Tree & \textbf{0.9991} & \textbf{0.9991} & \textbf{0.9991} & \textbf{0.9991} & \textbf{1.0000} 
        & 0.7507 & 0.7498 & 0.7551 & 0.7507 & 0.7504 \\ \hline
        \multicolumn{11}{|c|}{\textbf{Linear parametric machine learning models}} \\ \hline
        Logistic Regression & 0.7448 & 0.7445 & 0.7456 & 0.7448 & 0.8168 
        & 0.7252 & 0.7251 & 0.7271 & 0.7252 & 0.7985 \\
        Linear SVM & 0.7357 & 0.7355 & 0.7364 & 0.7357 & 0.7357 
        & 0.7212 & 0.7211 & 0.7233 & 0.7212 & 0.7219 \\ \hline
        \multicolumn{11}{|c|}{\textbf{Nonlinear parametric machine learning models}} \\ \hline
        Gaussian Process & 0.9954 & 0.9954 & 0.9955 & 0.9954 & 0.9992 
        & 0.8376 & 0.8376 & 0.8391 & 0.8376 & 0.9098 \\
        Neural Net & 0.9074 & 0.9074 & 0.9082 & 0.9074 & 0.9673 
        & 0.7872 & 0.7872 & 0.7888 & 0.7872 & 0.8622 \\
        RBF SVM & \textbf{0.9991} & \textbf{0.9991} & \textbf{0.9991} & \textbf{0.9991} & \textbf{0.9991} 
        & 0.6967 & 0.6662 & 0.8097 & 0.6967 & 0.6966 \\
        QDA & 0.6132 & 0.6011 & 0.6259 & 0.6132 & 0.6672 
        & 0.6031 & 0.5851 & 0.6169 & 0.6031 & 0.6606 \\ \hline
        \multicolumn{11}{|c|}{\textbf{Bayesian-based machine learning models}} \\ \hline
        Naive Bayes & 0.6079 & 0.5467 & 0.7347 & 0.6079 & 0.7030 
        & 0.6079 & 0.5474 & 0.7348 & 0.6079 & 0.6941 \\ \hline
    \end{tabular}%
    }
    \label{tab:performance}
\vspace{-10pt}
\end{table*}

\begin{figure}[!t]
\centerline{\includegraphics[width=0.5\columnwidth]{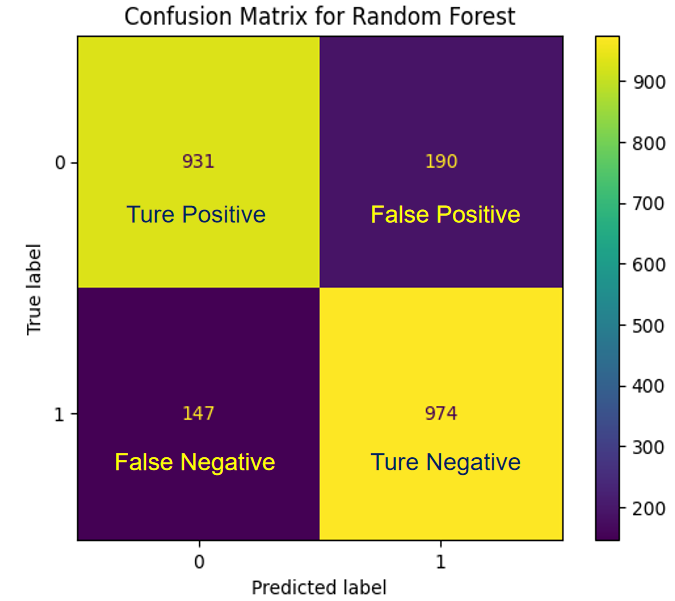}}
\caption{Confusion matrix of random forest model trained on oversampled and one-hot encoded data classified by V2 dosage.}
\label{fig2}

\end{figure}

In selecting models for this paper, a comprehensive approach is taken to comprise 17 binary classification techniques covering most types of machine learning models:
\begin{itemize}
    \item Ensemble machine learning models: gradient boosting classifier, hist gradient boosting classifier, adaBoost classifier, random forest, voting classifier soft, extra trees, stacking classifier, and catboost classifier.
    \item Nonparametric machine learning models: k-nearest neighbor and decision tree.
    \item Linear parametric machine learning models: logistic regression and linear svm.
    \item Nonlinear parametric machine learning models: quadratic discriminant analysis (QDA), neural network: multilayer perceptron classifier, RBF SVM and gaussian process.
    \item Bayesian-based machine learning models: Gaussian Naive Bayes
\end{itemize}

This selection is motivated by the need to capture various properties of the data, benchmark performance across different methodologies, and enhance robustness and generalization. By evaluating a wide range of models, we aim at identifying the best-performing model for our dataset.

During the model training phase, we employ 5-fold cross-validation to enhance the robustness and generalization capability of the models, providing a more reliable assessment of their performance. The trained models are evaluated across multiple dimensions, including Accuracy, F1 score,  Precision, Recall, and ROC-AUC. The performance of the models is presented in TABLE \ref{tab:performance}, where the Random Forest outperformed all other 16 models, achieving approximately 0.85 in various metrics. Consequently, we select the Random Forest model for the application of CFs techniques. Note that some evaluation metric values of the training set approach 1. However, this is attributed to the inherent nature of the small data, which is clean and devoid of anomalies or missing values. Additionally, the categorization of antidepressants is to some extent associated with the severity of depression, which in turn is measured by HAM-D scores. Hence, the model can readily acquire knowledge from the data.

To further illustrate the effectiveness of the Random Forest model, we present the confusion matrix in Fig. \ref{fig2}, which offers a detailed breakdown of the model's prediction performance. Specifically, the matrix shows that out of 1121 instances of class 0 (SSRIs), the model correctly classified 931 instances, resulting in 190 false positives. For class 1 (SNRIs), the model correctly classified 974 out of 1121 instances, with 147 false negatives. This visualization provides deeper awareness of the model's strengths by highlighting its ability to correctly classify instances and identifying the types of errors it makes.

\subsection{Sample Based Counterfactual Explanation}

\begin{figure*}[!htbp]
    \centering
    \begin{minipage}[b]{0.4\textwidth}
        \centering
        \includegraphics[width=\linewidth]{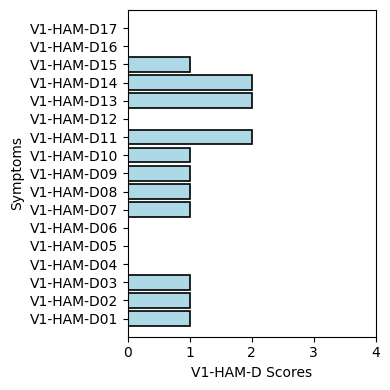}
        \\ \textbf{(a)} Original instance (class 0-SSRI)
        \label{fig3a}
    \end{minipage}%
    \hfill
    \begin{minipage}[b]{0.4\textwidth}
        \centering
        \includegraphics[width=\linewidth]{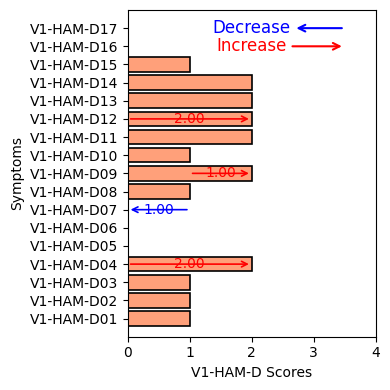}
        \\ \textbf{(b)} One counterfactual example (class 1-SNRI)
        \label{fig3b}
    \end{minipage}
    \caption{Sample-based counterfactual explanations on a random instance. This random instance is classified as 0 indicating SSRI medication is prescribed in the model prediction. The counterfactual example with a constrained counterfactual class of 1 is driven by changes in the V1-HAM-D scores, with the V1-HAM-D10 (psychic anxiety) score manually set to remain unchanged. The arrows indicate the direction and magnitude of change.}
    \label{fig3}
\end{figure*}

Following the selection of the Random Forest model, we apply CFs techniques to gain deeper perspectives on its decision-making process at personalized level.

Fig. \ref{fig3} illustrates how a counterfactual example is generated. The direction of the arrows indicates the change of HAM-D scores between the original instance and counterfactual example (i.e., left for a decrease, right for an increase), and the numbers on the arrows represent the magnitude of the change.

As mentioned earlier, real-life situations often impose numerous constraints, which may contribute to greater treatment resistance, making it challenging to achieve idealized symptom changes. Therefore, it is pivotal to consider these practical limitations in generating CFs. In light of this, we generate counterfactual example (Fig. \ref{fig3}(b)) for the original Instance (Fig. \ref{fig3}(a)), taking into account the realistic constraint that assuming the V1-HAM-D10 (psychic anxiety) is one of the symptoms that cannot feasibly be changed in reality over a short period. Consequently, the model makes adjustments, such as increasing V1-HAM-D12 by 2 units and V1-HAM-D4 by 2 units, respectively, to re-classify the instance as SNRI. These specific adjustments quantitatively elucidated how altering certain depressive symptoms can lead to different treatment recommendations, highlighting the causal relationship between symptom severity and medication categories. By examining this ‘what if' scenario with realistic constraints, we can see how the model adapts its recommendations, providing more nuanced and feasible treatment options.

\subsection{Local Feature Importance}

\begin{figure*}[!htbp]
    \centering
    \begin{minipage}[b]{0.35\textwidth}
        \centering
        \includegraphics[width=\linewidth]{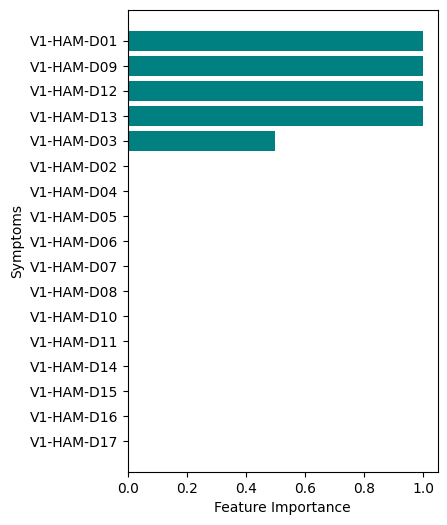}
        \\ \textbf{(a)} Local feature importance result on a random instance
        \label{fig4a}
    \end{minipage}%
    \hfill 
    \begin{minipage}[b]{0.35\textwidth}
        \centering
        \includegraphics[width=\linewidth]{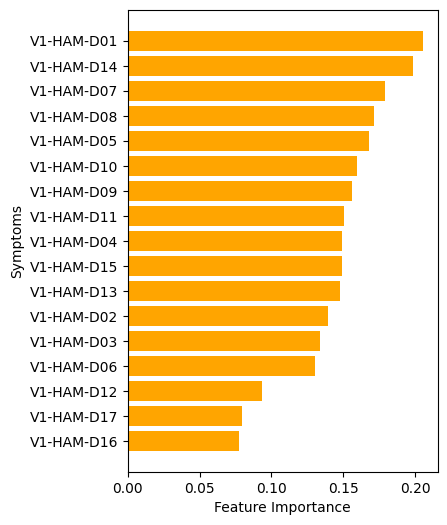}
        \\ \textbf{(b)} Global feature importance result on all instances
        \label{fig4b}
    \end{minipage}
    \caption{Feature importance results. (a) Local feature importance calculated from the minimum changes between the original input and 10 CFs. The most influential features contributing to the model prediction (medication selection) at the personalized level are V1-HAM-D01 (depressed mood), V1-HAM-D09 (psychomotor agitation), V1-HAM-D12 (loss of appetite), V1-HAM-D13 (tiredness/pain), and V1-HAM-D03 (suicidal thoughts or actions). (b) Global feature importance result on all instances, where the most influential feature contributing to the model prediction on medication selection at the population level is V1-HAM-D01 (depressed mood), while the least contributing one is V1-HAM-D16 (lack of insight).}
    \label{fig4}
\end{figure*}

Local feature importance also plays a critical role in understanding the decision-making process at personalized level. A feature, consistently changing during the generation of CFs for a given instance, is considered more influential in driving the model's prediction for that instance.

To calculate local feature importance, we assess the frequency of feature alterations during the generation of CFs for a specific instance. Since a single counterfactual example cannot capture the frequency of feature changes comprehensively, we generate 10 CFs for the original instance. By analyzing these CFs, we could quantify the frequency of feature changes and derive the local feature importance, as depicted in Fig. \ref{fig4}(a). Larger values of feature importance indicate a greater influence of the corresponding symptom on the prediction outcomes.

This concept of local feature importance can provide valuable assistance in offering personalized treatment recommendations based on individual patient symptoms. By identifying which symptoms have the top influence, such as V1-HAM-D01, V1-HAM-D09, V1-HAM-D12, V1-HAM-D13, and V1-HAM-D03, clinicians can distinguish the important factors affecting a specific patient, while recognizing that other symptoms may not exert a decisive influence on the individual's condition. Leveraging this impression, we can provide tailored treatments for specific cases, utilizing the identified key factors to address the patient's unique needs.

\subsection{Global Feature Importance}
Having presented local feature importance, it is imperative to explore the concept of global feature importance and its significance in model interpretation at the population level. While local feature importance explains the influence of symptoms on specific predictions of one sample (considered as personalized level), global feature importance offers a broader perspective by considering the overall impact of symptoms across the entire dataset (considered as population level).

Global feature importance is calculated by aggregating the local importance scores obtained from all CFs. 
To quantify global feature importance, we utilize the entire dataset and generate 10 CFs for each instance. By analyzing the aggregated results, we derive the global feature importance scores.

The Fig. \ref{fig4}(b) quantifies the specific degree of symptom impacts, V1-HAM-D01 (Depressed mood) is identified as the most important symptom in influencing model predictions, whereas V1-HAM-D16 (Weight loss) is shown as the least important symptom. 

\textcolor{black}{We then employ an expert-centered evaluation to validate the realism and clinical relevance of the generated CFs. This expert validation is crucial for ensuring the practical applicability of CFs, particularly in healthcare settings where clinical relevance and actionability are essential \cite{arrieta2020explainable}. Our global feature importance analysis supports the validity of these CFs, showing that symptoms e.g., "depressed mood" (HAM-D01) and "loss of sexual interest" (HAM-D14) are among the most influential in determining medication selection, while "weight loss" (HAM-D16) has minimal impact. This finding aligns with clinical understanding and underscores the importance of focusing on key symptoms that clinicians prioritize when making medication decisions \cite{carrozzino2020hamilton,rabinowitz2022consistency}.}

\section{Discussion}

We explore the application of counterfactual reasoning in the selection of depression medication, focusing on both personalized and population-level analyses. Our approach utilizes CFs to investigate the causal relationships between HAM-D symptoms and the prescribed selection of medications (SSRIs/SNRIs), representing a significant advancement in combining counterfactual reasoning with AI CDSSs.


A key strength of our approach is its ability to generate personalized insights through local feature importance analysis. By quantifying the relative influence of individual HAM-D symptoms on medication decisions for each patient, our approach assists clinicians to develop tailored treatment strategies that precisely target the most pertinent factors driving a specific case. Notably, our paper accounts for real-world constraints by generating counterfactual scenarios that accommodate practical limitations, such as the inability to modify certain symptoms due to patient-specific factors. Complementing this personalized perspective, our global feature importance analysis provides a comprehensive population level view on the relative significance of various depressive symptoms in guiding medication selections across the entire dataset. By bridging individual and population analyses, our approach offers a holistic understanding of these complex relationships.


\textcolor{black}{It is well known that the overwhelming majority of choices in pharmacological management for depression remain SSRI’s and to a lesser extent SNRI’s. As such, investigating for factors that predict better outcome in these classes is sorely needed to help the pharmacologist match the patient to the appropriate medication. Furthermore, future datasets will include novel pharmacotherapeutics and as such, will be able to include these new agents as a comparator. Still, providing a baseline of how to separate what are appropriate choices in terms of SSRIs or SNRIs will allow for a comparison in the future of novel treatments to further help understand the pharmacodynamics of these new agents and specifically where they fit in the algorithm. Our hope is that the results of this work will be built off for future analyses that can ultimately come together to give a comprehensive understanding of who responds best (or worst) to which medication based on the causal relationship between symptoms and medication.  }

\textcolor{black}{Despite the valuable insights presented in our study, there are some limitations. The dataset derived from clinical trials of specific medications may not fully represent all patient groups, and future studies should include a more diverse patient demographic and a broader range of antidepressant classes. Additionally, the computational complexity of our algorithm is relatively high, suggesting the need for further optimization to enhance its practicality in clinical settings.}

\section{Conclusion}
In this paper, we present a counterfactual reasoning approach based on explainable counterfactual reasoning to investigate the causal relationship between the HAM-D scales and the categories of anti-depressant medication prescribed by clinicians. Our method employs counterfactual scenarios to simulate causal relationships, elucidating not only the causal relationship but also deriving feature importance from the disparities between CFs and the original instances. In this case we found that “depressed mood” (HAM-D01) and “loss of sexual interest” (HAM-D14) were features that were most influential in determining medication selection, while “weight loss” (HAM-D16) was least important. This approach offers valuable AI CDSSs by providing interpretations into the causal relationships and feature importance underlying clinical decision-making process. By addressing both personalized and population level analyses, our approach enhances the ability to tailor treatments to individual patient needs while also understanding broader trends and patterns within the population.


\section*{Declarations}

\subsection*{Abbreviations}
As a reference for readers, all abbreviations used in this manuscript are listed in Table~\ref{tab:abbreviations}.

\begin{table}[!htbp]
\centering
\caption{Abbreviations}
\label{tab:abbreviations}
\begin{tabular}{ll}
\hline
Abbreviation & Full term \\
\hline
CFs  & Counterfactual Explanations \\
CDSS & Clinical Decision Support System \\
HAM-D & Hamilton Rating Scale for Depression \\
MDD  & Major Depressive Disorder \\
SNRI & Serotonin-Norepinephrine Reuptake Inhibitor \\
SSRI & Selective Serotonin Reuptake Inhibitor \\
EHRs & Electronic Health Records \\
ML   & Machine Learning \\
XAI  & eXplainable Artificial Intelligence \\
DICE & Diverse Counterfactual Explanations\\
\hline
\end{tabular}
\end{table}



\subsection*{Ethics approval and consent to participate}
The data used in this analysis was anonymized data from a database of multiple clinical trials provided by a pharmaceutical company. Data were collected across sites throughout the United States and Canada in alignment with FDA and Health Canada regulations, respectively, and were monitored by the pharmaceutical company. Each original clinical trial was approved by the relevant institutional review boards or ethics committees, as disclosed in the primary disclosures of the clinical trial data on \url{http://clinicaltrials.gov} and within the text of the corresponding published papers. Informed consent was obtained from each patient or their legal representative and the investigator before any study-related procedures commenced, in accordance with the requirements of the FDA, Health Canada, and other applicable regulations. The study was conducted in compliance with the principles of the Declaration of Helsinki.
Clinical trial number: not applicable.

\subsection*{Consent for publication}
Not applicable.

\subsection*{Availability of data and materials}
We cannot provide a direct link to the datasets used in this analysis because we were provided access by the pharmaceutical company and cannot download or transfer this anonymized data.

\subsection*{Competing Interests}
The authors declare that they have no competing interests.

\subsection*{Funding}
The research presented in this work was supported by the National Science Foundation NSF under grant number CRII:SCH:2437784 and by the Natural Sciences and Engineering Research Council of Canada NSERC through Discovery Grant RGPIN-2024-05683.

\subsection*{Authors' contributions}
Xinyu Qin conceived the study design, implemented the algorithms, ran the experiments, analysed the data, and drafted the manuscript. 
Mark H. Chignell, Alexandria Greifenberger, Sachinthya Lokuge, Elssa Toumeh, Tia Sternat, and Martin Katzman provided clinical background knowledge and dataset, interpreted the findings, and revised the manuscript. 
Lu Wang supervised the project, guided the methodology, and critically revised the manuscript. 
All authors read and approved the final manuscript.

\subsection*{Acknowledgements}
Thanks are extended to Eli Lilly and Company for providing access to the clinical trial data, and to all investigators and participants of the original studies.

\bibliography{sn-bibliography}

\begin{thebibliography}{10}
\providecommand{\doi}[1]{\url{https://doi.org/#1}}
\bibcommenthead

\bibitem[\protect\citeauthoryear{Sinyor et~al.}{2016}]{sinyor2016suicide}
Sinyor M, Tan LPL, Schaffer A, Gallagher D, Shulman K.
\newblock Suicide in the oldest old: an observational study and cluster analysis.
\newblock International journal of geriatric psychiatry. 2016;31(1):33--40.

\bibitem[\protect\citeauthoryear{Carrozzino et~al.}{2020}]{carrozzino2020hamilton}
Carrozzino D, Patierno C, Fava GA, Guidi J.
\newblock The Hamilton rating scales for depression: a critical review of clinimetric properties of different versions.
\newblock Psychotherapy and psychosomatics. 2020;89(3):133--150.

\bibitem[\protect\citeauthoryear{McIntyre et~al.}{2002}]{mcintyre2002assessing}
McIntyre R, Kennedy S, Bagby RM, Bakish D.
\newblock Assessing full remission.
\newblock Journal of Psychiatry and Neuroscience. 2002;27(4):235--239.

\bibitem[\protect\citeauthoryear{Hamilton}{1960}]{hamilton1960rating}
Hamilton M.
\newblock A rating scale for depression.
\newblock Journal of neurology, neurosurgery, and psychiatry. 1960;23(1):56.

\bibitem[\protect\citeauthoryear{Hamilton}{1967}]{hamilton1967development}
Hamilton M.
\newblock Development of a rating scale for primary depressive illness.
\newblock British journal of social and clinical psychology. 1967;6(4):278--296.

\bibitem[\protect\citeauthoryear{Nixon et~al.}{2020}]{nixon2020bi}
Nixon N, Guo B, Garland A, Kaylor-Hughes C, Nixon E, Morriss R.
\newblock The bi-factor structure of the 17-item Hamilton Depression Rating Scale in persistent major depression; dimensional measurement of outcome.
\newblock PloS one. 2020;15(10):e0241370.

\bibitem[\protect\citeauthoryear{Papakostas et~al.}{2007}]{papakostas2007augmentation}
Papakostas GI, Shelton RC, Smith J, Fava M.
\newblock Augmentation of antidepressants with atypical antipsychotic medications for treatment-resistant major depressive disorder: a meta-analysis.
\newblock Journal of Clinical Psychiatry. 2007;68(6):826--831.

\bibitem[\protect\citeauthoryear{Kennedy et~al.}{2016}]{kennedy2016canadian}
Kennedy SH, Lam RW, McIntyre RS, Tourjman SV, Bhat V, Blier P, et~al.
\newblock Canadian Network for Mood and Anxiety Treatments (CANMAT) 2016 clinical guidelines for the management of adults with major depressive disorder: section 3. Pharmacological treatments.
\newblock The Canadian Journal of Psychiatry. 2016;61(9):540--560.

\bibitem[\protect\citeauthoryear{Thase et~al.}{2001}]{thase2001remission}
Thase ME, Entsuah AR, Rudolph RL.
\newblock Remission rates during treatment with venlafaxine or selective serotonin reuptake inhibitors.
\newblock The British journal of psychiatry. 2001;178(3):234--241.

\bibitem[\protect\citeauthoryear{Shelton}{2019}]{shelton2019serotonin}
Shelton RC.
\newblock Serotonin and norepinephrine reuptake inhibitors.
\newblock Antidepressants: From Biogenic Amines to New Mechanisms of Action. 2019;p. 145--180.

\bibitem[\protect\citeauthoryear{Garcia-Toro et~al.}{2012}]{garcia2012treatment}
Garcia-Toro M, Medina E, Galan JL, Gonzalez MA, Maurino J.
\newblock Treatment patterns in major depressive disorder after an inadequate response to first-line antidepressant treatment.
\newblock BMC psychiatry. 2012;12:1--6.

\bibitem[\protect\citeauthoryear{Gianfrancesco et~al.}{2018}]{gianfrancesco2018potential}
Gianfrancesco MA, Tamang S, Yazdany J, Schmajuk G.
\newblock Potential biases in machine learning algorithms using electronic health record data.
\newblock JAMA internal medicine. 2018;178(11):1544--1547.

\bibitem[\protect\citeauthoryear{Arrieta et~al.}{2020}]{arrieta2020explainable}
Arrieta AB, D{\'\i}az-Rodr{\'\i}guez N, Del~Ser J, Bennetot A, Tabik S, Barbado A, et~al.
\newblock Explainable Artificial Intelligence (XAI): Concepts, taxonomies, opportunities and challenges toward responsible AI.
\newblock Information fusion. 2020;58:82--115.

\bibitem[\protect\citeauthoryear{Holzinger et~al.}{2019}]{holzinger2019causability}
Holzinger A, Langs G, Denk H, Zatloukal K, M{\"u}ller H.
\newblock Causability and explainability of artificial intelligence in medicine.
\newblock Wiley Interdisciplinary Reviews: Data Mining and Knowledge Discovery. 2019;9(4):e1312.

\bibitem[\protect\citeauthoryear{Roese and Olson}{2014}]{roese2014might}
Roese NJ, Olson JM.
\newblock What might have been: The social psychology of counterfactual thinking.
\newblock Psychology Press; 2014.

\bibitem[\protect\citeauthoryear{Hern{\'a}n and Robins}{2010}]{hernan2010causal}
Hern{\'a}n MA, Robins JM.: Causal inference.
\newblock CRC Boca Raton, FL.

\bibitem[\protect\citeauthoryear{Kim et~al.}{2016}]{kim2016examples}
Kim B, Khanna R, Koyejo OO.
\newblock Examples are not enough, learn to criticize! criticism for interpretability.
\newblock Advances in neural information processing systems. 2016;29.

\bibitem[\protect\citeauthoryear{Wachter et~al.}{2017}]{wachter2017counterfactual}
Wachter S, Mittelstadt B, Russell C.
\newblock Counterfactual explanations without opening the black box: Automated decisions and the GDPR.
\newblock Harv JL \& Tech. 2017;31:841.

\bibitem[\protect\citeauthoryear{Poyiadzi et~al.}{2020}]{poyiadzi2020face}
Poyiadzi R, Sokol K, Santos-Rodriguez R, De~Bie T, Flach P.
\newblock FACE: feasible and actionable counterfactual explanations.
\newblock In: Proceedings of the AAAI/ACM Conference on AI, Ethics, and Society; 2020. p. 344--350.

\bibitem[\protect\citeauthoryear{Laugel et~al.}{2017}]{laugel2017inverse}
Laugel T, Lesot MJ, Marsala C, Renard X, Detyniecki M.
\newblock Inverse classification for comparison-based interpretability in machine learning.
\newblock arXiv preprint arXiv:171208443. 2017;.

\bibitem[\protect\citeauthoryear{Dandl et~al.}{2020}]{dandl2020multi}
Dandl S, Molnar C, Binder M, Bischl B.
\newblock Multi-objective counterfactual explanations.
\newblock In: International Conference on Parallel Problem Solving from Nature. Springer; 2020. p. 448--469.

\bibitem[\protect\citeauthoryear{Mothilal et~al.}{2020}]{mothilal2020explaining}
Mothilal RK, Sharma A, Tan C.
\newblock Explaining machine learning classifiers through diverse counterfactual explanations.
\newblock In: Proceedings of the 2020 conference on fairness, accountability, and transparency; 2020. p. 607--617.

\bibitem[\protect\citeauthoryear{Ribera and Lapedriza}{2019}]{ribera2019can}
Ribera M, Lapedriza A.
\newblock Can we do better explanations? A proposal of user-centered explainable AI.
\newblock CEUR Workshop Proceedings; 2019. .

\bibitem[\protect\citeauthoryear{Ribeiro et~al.}{2016}]{ribeiro2016should}
Ribeiro MT, Singh S, Guestrin C.
\newblock " Why should i trust you?" Explaining the predictions of any classifier.
\newblock In: Proceedings of the 22nd ACM SIGKDD international conference on knowledge discovery and data mining; 2016. p. 1135--1144.

\bibitem[\protect\citeauthoryear{Lundberg and Lee}{2017}]{lundberg2017unified}
Lundberg SM, Lee SI.
\newblock A unified approach to interpreting model predictions.
\newblock Advances in neural information processing systems. 2017;30.

\bibitem[\protect\citeauthoryear{Craven and Shavlik}{1995}]{craven1995extracting}
Craven M, Shavlik J.
\newblock Extracting tree-structured representations of trained networks.
\newblock Advances in neural information processing systems. 1995;8.

\bibitem[\protect\citeauthoryear{Kommiya~Mothilal et~al.}{2021}]{kommiya2021towards}
Kommiya~Mothilal R, Mahajan D, Tan C, Sharma A.
\newblock Towards unifying feature attribution and counterfactual explanations: Different means to the same end.
\newblock In: Proceedings of the 2021 AAAI/ACM Conference on AI, Ethics, and Society; 2021. p. 652--663.

\bibitem[\protect\citeauthoryear{Ali et~al.}{2023}]{ali2023enlightening}
Ali S, Akhlaq F, Imran AS, Kastrati Z, Daudpota SM, Moosa M.
\newblock The enlightening role of explainable artificial intelligence in medical \& healthcare domains: A systematic literature review.
\newblock Computers in Biology and Medicine. 2023;p. 107555.

\bibitem[\protect\citeauthoryear{Kulesza et~al.}{2012}]{kulesza2012determinantal}
Kulesza A, Taskar B, et~al.
\newblock Determinantal point processes for machine learning.
\newblock Foundations and Trends{\textregistered} in Machine Learning. 2012;5(2--3):123--286.

\bibitem[\protect\citeauthoryear{Nitesh}{2002}]{nitesh2002smote}
Nitesh VC.
\newblock SMOTE: synthetic minority over-sampling technique.
\newblock J Artif Intell Res. 2002;16(1):321.

\bibitem[\protect\citeauthoryear{Rabinowitz et~al.}{2022}]{rabinowitz2022consistency}
Rabinowitz J, Williams JB, Anderson A, Fu DJ, Hefting N, Kadriu B, et~al.
\newblock Consistency checks to improve measurement with the Hamilton Rating Scale for Depression (HAM-D).
\newblock Journal of Affective Disorders. 2022;302:273--279.

\end{thebibliography}
\end{document}